\documentclass[sigconf]{aamas}

\usepackage{hyperref}
\usepackage{url}
\usepackage{algorithm, algorithmic}

\usepackage{booktabs}
\usepackage{amsmath, mathtools, amsthm}
\usepackage[capitalize,noabbrev]{cleveref}
\usepackage{multirow}
\usepackage{color}

\usepackage{balance} 

\doi{GZIN7614}

\makeatletter
\gdef\@copyrightpermission{
  \begin{minipage}{0.2\columnwidth}
   \href{https://creativecommons.org/licenses/by/4.0/}{\includegraphics[width=0.90\textwidth]{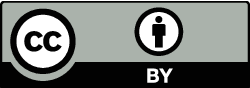}}
  \end{minipage}\hfill
  \begin{minipage}{0.8\columnwidth}
   \href{https://creativecommons.org/licenses/by/4.0/}{This work is licensed under a Creative Commons Attribution International 4.0 License.}
  \end{minipage}
  \vspace{5pt}
}
\makeatother

\setcopyright{ifaamas}
\acmConference[AAMAS '26]{Proc.\@ of the 25th International Conference
on Autonomous Agents and Multiagent Systems (AAMAS 2026)}{May 25 -- 29, 2026}
{Paphos, Cyprus}{C.~Amato, L.~Dennis, V.~Mascardi, J.~Thangarajah (eds.)}
\copyrightyear{2026}
\acmYear{2026}
\acmDOI{}
\acmPrice{}
\acmISBN{}

\title[AAMAS-2026 Formatting Instructions]{PIQL: Projective Implicit Q-Learning with Support Constraint for Offline Reinforcement Learning}

\author{Xinchen Han}
\affiliation{
  \institution{SAMOVAR, Institut Polytechnique de Paris}
  \city{Paris}
  \country{France}}
\email{xinchen.han@telecom-sudparis.eu}

\author{Hossam Afifi}
\affiliation{
  \institution{SAMOVAR, Institut Polytechnique de Paris}
  \city{Paris}
  \country{France}}
\email{hossam.afifi@telecom-sudparis.eu}

\author{Michel Marot}
\affiliation{
  \institution{SAMOVAR, Institut Polytechnique de Paris}
  \city{Paris}
  \country{France}}
\email{michel.marot@telecom-sudparis.eu}

\begin{abstract}
Offline Reinforcement Learning (RL) faces a fundamental challenge of extrapolation errors caused by out-of-distribution (OOD) actions. Implicit Q-Learning (IQL) employs expectile regression to achieve in-sample learning. Nevertheless, IQL relies on a fixed expectile hyperparameter and a density-based policy improvement method, both of which impede its adaptability and performance. In this paper, we propose Projective IQL (PIQL), a projective variant of IQL enhanced with a support constraint. In the policy evaluation stage, PIQL substitutes the fixed expectile hyperparameter with a projection-based parameter and extends the one-step value estimation to a multi-step formulation. In the policy improvement stage, PIQL adopts a support constraint instead of a density constraint, ensuring closer alignment with the policy evaluation. Theoretically, we demonstrate that PIQL maintains the expectile regression and in-sample learning framework, guarantees monotonic policy improvement, and introduces a progressively more rigorous criterion for advantageous actions. Experiments on D4RL and NeoRL2 benchmarks demonstrate robust gains across diverse domains, achieving state-of-the-art performance overall.
\end{abstract}


\newcommand{\BibTeX}{\rm B\kern-.05em{\sc i\kern-.025em b}\kern-.08em\TeX}
\theoremstyle{plain}
\newtheorem{theorem}{Theorem}

\newtheorem{lemma}{Lemma}

\theoremstyle{definition}

\theoremstyle{remark}

\begin{document}


\pagestyle{fancy}
\fancyhead{}


\maketitle 

\section{Introduction}
Offline RL aims to learn policies from static datasets collected by unknown behavior policy, thereby avoiding costly or unsafe online interactions with the environment~\cite{Survey_Prudencio, Survey_Levine}. This data-driven paradigm holds strong promise for real-world applications~\cite{LeakyPPO}. However, removing continuous interaction also introduces substantial challenges.
A central difficulty is extrapolation error~\cite{BCQ}, which arises from OOD actions that are not represented in the dataset. Such errors can be amplified through bootstrapped Bellman backups and may distort value estimates for a wide range of state–action pairs, leading to degraded policy performance. To address this challenge, existing research has elaborated offline RL algorithms focusing on policy improvement and policy evaluation, respectively.

In the policy improvement stage, policy constraint methods encourage the learned policy to remain close to the behavior policy. Behavior cloning (BC)~\cite{BC} imitates dataset actions via supervised learning, while weighted BC (wBC) methods~\cite{RWR, AWR, AWAC, CRR, Cos-IQL, ACPO} generalize BC by reweighting samples according to different criteria. Many wBC-style objectives can be interpreted as imposing a KL-style~\cite{KLdiv} regularization, which can be categorized into density constraint methods. In contrast, support constraint methods~\cite{BEAR, SPOT, STR, SVR} apply looser constraints that encourage the learned policy to remain within the support of the behavior policy while permitting more substantial improvement.

In the policy evaluation stage, conservative value learning methods mitigate overestimation issues induced by OOD actions~\cite{CQL, CEN, IQL, AEM}. Especially, IQL~\cite{IQL} employs expectile regression to approximate an upper expectile of the value distribution instead of relying on the traditional maximum operator in Q-Learning. However, IQL has two notable limitations. First, IQL depends on a fixed expectile hyperparameter, which often requires dataset-specific tuning to achieve good performance. Second, IQL performs essentially one-step method~\cite{Onestep}, limiting its extensibility to more powerful multi-step policy improvement framework. 

To overcome these challenges, we propose Projective IQL (PIQL), a projective variant of IQL enhanced with a support constraint. PIQL introduces two crucial innovations. First, PIQL replaces the hyperparameter $\tau$ with projection-based adaptive parameter $\tau_{\text{proj}}(a|s)$. This parameter $\tau_{\text{proj}}(a|s)$ is computed by projecting the behavior policy onto the learned policy, thereby eliminating the need for dataset-specific tuning. This formulation preserves the in-sample learning and expectile regression framework of IQL, while naturally generalizing one-step value estimation to a multi-step formulation. Second, PIQL incorporates support constraint aligned with the policy evaluation stage, achieving more effective policy improvement while mitigating extrapolation error.

Theoretically, we demonstrate that PIQL can be interpreted as a multi-step offline RL method while retaining the expectile-regression objective and the in-sample learning paradigm. Moreover, PIQL is provably guaranteed to yield monotonic policy improvement and to enforce a progressively more rigorous criterion for identifying and selecting advantageous actions.

Experimentally, PIQL achieves state-of-the-art performance on D4RL~\cite{D4RL}, with particularly strong results on challenging ``stitching'' navigation tasks that require composing sub-trajectories over long horizons. We further evaluate PIQL on the recent NeoRL2~\cite{NeoRL2} benchmarks, where it consistently surpasses the strongest baselines.


\section{Related Work} \label{Sec-RelatedWork}

To more effectively compare the advantages of PIQL, we broadly classify existing offline RL approaches into two categories: conservative value methods and policy constraint methods.

Conservative value methods aim to mitigate the overestimation of OOD actions by penalizing overly optimistic value estimates. CQL \cite{CQL} introduces penalties on the $Q$-function for OOD actions, ensuring policy performance while staying close to the behavior policy. Extensions like ACL-QL \cite{ACL-QL} dynamically adjusts penalty term to reduce over-conservatism, and CSVE \cite{CSVE} penalizes state value function estimates across the dataset to enhance generalization. Despite their efficacy in avoiding overestimation, these approaches still face OOD action risk. In-sample learning approaches such as Onestep \cite{Onestep}, IQL \cite{IQL}, and IVR \cite{SQL} eliminate the need to estimate $Q$-values for OOD actions entirely. Onestep demonstrates strong results with one-step policy improvement but relies on well-covered datasets. IVR introduces an implicit value regularization framework, but Sparse Q-Learning within IVR relies on a restrictive assumption to simplify its optimization objective. IQL leverages the expectile regression framework for value learning, but its hyperparameter $\tau$ requires careful tuning, which is time-consuming.

Policy constraint methods enforce closeness between the learned policy and the behavior policy using various metrics such as KL divergence \cite{AWAC, BRAC+}, regularization terms \cite{TD3+BC}, or CVAE model \cite{BCQ, PLAS, LAPO, ELAPSE}, etc. While these density-based methods perform well under moderate suboptimality, they struggle when the dataset predominantly consists of poor-quality data. Support-based approaches relax density constraints to allow greater optimization flexibility. BEAR \cite{BEAR} keep the policy within the support of behavior policy by Maximum Mean Discrepancy (MMD). By contrast, SPOT \cite{SPOT} adopts a VAE-based density estimator to explicitly model the support set of behavior policy. Similarly, DASCO \cite{DASCO} employs discriminator scores in Generative Adversarial Networks (GANs). Although these methods achieve strong empirical performance, they lack rigorous policy improvement guarantees \cite{STR}. 

In contrast to conservative value methods, PIQL preserves the in-sample learning benefits of IQL while extending it into a multi-step formulation through a projection-based parameter. Unlike density-based policy constraint approaches, PIQL belongs to the family of support-constraint methods, thereby enabling greater policy optimization flexibility. Moreover, although PIQL adopts a policy improvement scheme similar to STR \cite{STR}, PIQL further enforces increasingly rigorous criteria for selecting advantageous actions from the dataset, thereby ensuring steady policy improvement.

\section{Preliminaries} \label{Sec-Preliminaries}

\subsection{Offline RL}

Conventionally, the RL problem is defined as a Markov Decision Process (MDP) \cite{SuttonRL}, specified by a tuple $\mathcal{M} = \langle \mathcal{S}, \mathcal{A}, \mathcal{T}, r, \gamma \rangle$, where $\mathcal{S}$ denotes the state space, $\mathcal{A}$ is the action space, $\mathcal{T}$ represents the transition function, $r$ is the reward function, and $\gamma \in [0,1)$ denotes the discount factor.

For policy $\pi(a|s)$, the $Q$-function $Q^{\pi}(s, a)$ is defined as the expectation of cumulative discounted future rewards,
\begin{gather} 
\begin{aligned}
Q^{\pi}(s, a) = \mathop{\mathbb{E}}_{\substack{a_t \sim \pi(\cdot|s_t),\\s_{t+1} \sim \mathcal{T} (\cdot |s_{t}, a_{t})}} \left[ \sum_{t=0}^{\infty} \gamma^t r(s_t, a_t) \Big| s_0=s, a_0=a \right].
\label{Eq-Q-function}
\end{aligned}
\end{gather}

The value function $V^{\pi}(s) = \mathbb{E}_{a \sim \pi} \left[ Q^{\pi}(s, a)\right]$, and the advantage function $A^{\pi}(s, a) = Q^{\pi}(s, a) - V^{\pi}(s)$. The $Q$-function is updated by minimizing the Temporal Difference (TD) error, shown in the following Eq.~\eqref{Eq-TD-error},
\begin{gather} 
\begin{aligned}
L_{\text{TD}}(\theta) = \mathop{\mathbb{E}}_{\substack{(s, a, s')\sim \mathcal{B}}} \left[ [r(s,a) + \gamma \max_{a'} Q_{\hat{\theta}}(s',a') - Q_\theta(s,a)]^2 \right].
\label{Eq-TD-error}
\end{aligned}
\end{gather}

When the state-action space is large, the $Q$-function is approximated by neural networks $Q_\theta(s, a)$ and parameterized by $\theta$, and the $Q_{\hat{\theta}}(s, a)$ represents the target-$Q$ networks whose parameters $\hat{\theta}$ are updated via Polyak averaging approach.

In the offline RL setting, the replay buffer $\mathcal{B}$ is replaced by the static dataset $\mathcal{D}$, and the agent is trained solely on $\mathcal{D}$ without any interaction with the environment for further data collection. If an OOD action $a' = \mathop{\arg\max}_{a'} Q(s',a')$ is chosen but is absent from the dataset $a' \notin \mathcal{D}$, the estimated value $Q(s', a')$ becomes arbitrary and unreliable. Such pathological estimates can destabilize the bootstrapping process. 

\subsection{Support Constraint}
The support-constrained policy class $\Pi$ is defined as 
\begin{gather} 
\begin{aligned}
\Pi = \{\pi| \pi(a|s)=0 \; \text{whenever} \; \pi_{\beta}(a|s)=0\},
\label{Eq-Support-constrained}
\end{aligned}
\end{gather}
where $\pi_{\beta}$ is the behavior policy used to build datasets.

The policy optimization objective of wBC methods are summarized in \cite{STR}, shown as follows,
\begin{gather} 
\begin{aligned}
J_\pi^{\text{wBC}}(\phi) = \mathop{\max}_{\phi} \mathop{\mathbb{E}}_{\substack{s \sim \mathcal{D}, \\a \sim \pi_{\text{base}}}} \left[ f(s, a; \pi_{\text{pe}}) \log \pi_\phi(a|s) \right],
\label{Eq-wBC-Pi}
\end{aligned}
\end{gather}
where $\pi_{\text{base}}$ denotes the sampling policy, and $\pi_{\text{pe}}$ denotes the evaluation policy. $f(s, a; \pi_{\text{pe}}) = \frac{1}{Z(s)}\exp\left(\frac{A^{\pi_{\text{pe}}}(s, a)}{\lambda} \right)$, $\lambda$ is a temperature parameter and $Z(s) = \int_{a} \pi_{\text{base}}(a|s) \exp\left(\frac{A^{\pi_{\text{pe}}}(s, a)}{\lambda} \right) da.$ But $Z(s)$ is usually omitted during training, because errors in the estimation of $Z(s)$ caused more harm than the benefit the method derived from estimating this value, thereby $f(s, a; \pi_{\text{pe}}) = \exp\left(\frac{A^{\pi_{\text{pe}}}(s, a)}{\lambda} \right)$.

Since $f(s, a; \pi_{\text{pe}}) > 0$, the policy update of wBC is an equal-support update, i.e. if we initialize $\pi_{\text{base}}$ as $\pi_{\beta}$ and choose the current policy $\pi_{\phi}$ to be $\pi_{\text{base}}$, by recursive improvement, $\pi_{\phi}$ is still within the support of $\pi_{\beta}$. Based on the key observation above, STR \cite{STR} proposes a support constraint policy improvement method with $\pi_{\text{base}} = \pi_{\phi}, \pi_{\text{pe}} = \pi_{\phi}$ and adopts the Importance Sampling (IS) technique,
\begin{gather} 
\begin{aligned}
J_\pi^{\text{STR}}(\phi) = \mathop{\max}_{\phi} \mathop{\mathbb{E}}_{(s,a) \sim \mathcal{D}} \left[ \frac{\Bar{\pi}_{\phi}(a|s)}{\pi_{\beta}(a|s)} f(s, a; \Bar{\pi}_{\phi}) \log \pi_\phi(a|s) \right],
\label{Eq-STR-Pi}
\end{aligned}
\end{gather}
where $\Bar{\pi}_{\phi}(a|s)$ is the learned policy with detached of gradient.

\subsection{IQL}

IQL utilizes expectile regression to modify the policy evaluation objective in Eq.~\eqref{Eq-TD-error}. The maximum operator is replaced by an upper expectile. In addition, to mitigate the stochasticity introduced by environment dynamics $s' \sim p(\cdot | s, a)$, IQL employs a separate value function that estimates an expectile solely with respect to the action distribution. Hence, the update of value function and the $Q$-function in IQL during policy evaluation are illustrated below:
\begin{gather} 
\begin{aligned}
L_V^{\text{IQL}}(\psi) = \mathop{\mathbb{E}}_{(s,a) \sim \mathcal{D}}\left[ L_2^\tau(Q_{\hat{\theta}}(s,a) - V_\psi(s)) \right],
\label{Eq-IQL-V}
\end{aligned}
\end{gather}
\begin{gather} 
\begin{aligned}
L_Q^{\text{IQL}}(\theta) = \mathop{\mathbb{E}}_{(s,a,s') \sim \mathcal{D}}\left[ (r(s,a) + \gamma V_\psi(s') - Q_{\theta}(s,a))^2 \right],
\label{Eq-IQL-Q}
\end{aligned}
\end{gather}
where $L_2^\tau(u) = |\tau-\mathbb{I}(u < 0)|u^2$, the $\tau$ expectile of some random variable is defined as a solution to the asymmetric least squares problem, and $\tau \in (0, 1)$.

IQL extract policy using the wBC method. The objective of policy improvement in IQL is defined as follows, 
\begin{gather} 
\begin{aligned}
J_\pi^{\text{IQL}}(\phi) = \mathop{\max}_{\phi} \mathop{\mathbb{E}}_{(s, a) \sim \mathcal{D}} \left[ f(s, a; \pi_{\beta}) \log \pi_\phi(a|s) \right].
\label{Eq-IQL-Pi}
\end{aligned}
\end{gather}

Following the definition of one-step algorithms \cite{Onestep}, IQL can be classified as a one-step algorithm since the actions used for policy evaluation are drawn directly from the behavior policy, i.e., $a \sim \pi_{\beta} \; \text{in} \; f(s, a; \pi_{\beta})$. Furthermore, the actions for policy improvement are also restricted to the dataset $a \sim \mathcal{D}$, which allows IQL to benefit from in-sample learning and thus avoid OOD risk. However, this property simultaneously limits its potential for policy improvement.


\section{PIQL} \label{Sec-PIQL}
In this section, we provide a detailed description of PIQL, including policy evaluation, policy improvement, and practical implementation.

\subsection{Policy Evaluation in PIQL}
For the sake of simplicity, we follow the notation in IQL \cite{IQL} to introduce PIQL. Let $\mathbb{E}_{x \sim X}^{\tau}[x]$ be a $\tau^{th}$ expectile of $X$, e.g. $\mathbb{E}^{0.5}$ is the standard expectation. Then the recursive relationship between $V_{\tau}(s)$ and $Q_{\tau}(s,a)$ are shown as follows,
\begin{gather} 
\begin{aligned}
V_{\tau}(s) = \mathbb{E}^{\tau}_{a} \left[Q_{\tau}(s, a)\right].
\label{Eq-IQL-V-tau}
\end{aligned}
\end{gather}
\begin{gather} 
\begin{aligned}
Q_{\tau}(s, a) = r(s, a) + \gamma \mathop{\mathbb{E}}_{\substack{s' \sim \mathcal{T}(\cdot |s, a)}} \left[V_{\tau}(s')\right].
\label{Eq-IQL-Q-tau}
\end{aligned}
\end{gather}

\begin{lemma} \label{lemma1}
For all $s$, $\tau_1$ and $\tau_2$ such that $\tau_1 < \tau_2$ we get
$$
V_{\tau_1}(s) \leq V_{\tau_2}(s).
$$
\end{lemma}

Based on Lemma~\ref{lemma1}, a larger $\tau$ leads to a more optimistic policy evaluation, with the $Q$-function update approaching the maximum operator in Q-learning as $\tau \rightarrow 1$. Conversely, a smaller $\tau$ yields a more conservative policy evaluation. This observation highlights the role of $\tau$ as a trade-off parameter that balances optimism and conservatism in policy evaluation, and thus has a critical influence on the overall performance of IQL.

\begin{figure*}[!htbp] 
\centerline{\includegraphics[width=1.0\textwidth]{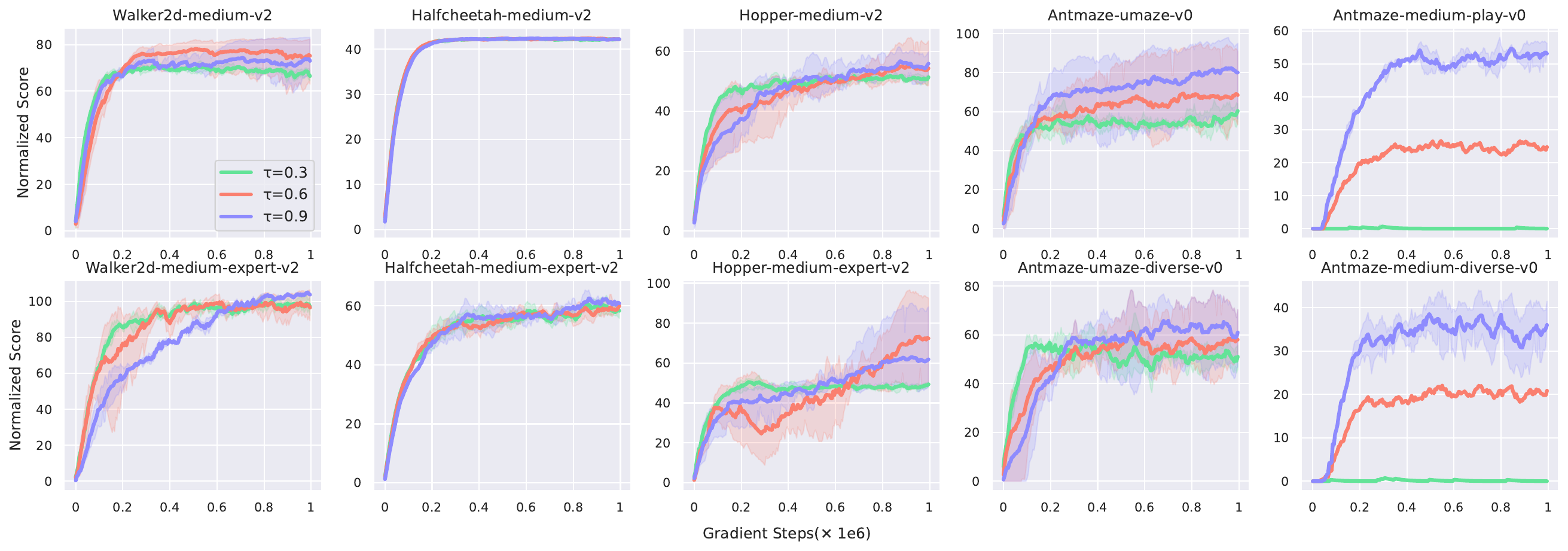}}
\caption{Effect of the Parameter $\tau \in \{0.3, 0.6, 0.9\}$ on IQL Performance across $10$ Datasets.}
\label{Fig-different-tau}
\end{figure*}

To illustrate the significance of the parameter $\tau$, we evaluate the performance of IQL under three settings, $\tau \in \{0.3, 0.6, 0.9\}$, across $10$ D4RL datasets. The results, presented in Fig.~\ref{Fig-different-tau}, show that the performance of IQL varies considerably with different values of $\tau$ on the same dataset. This indicates that arbitrarily choosing $\tau$ may lead to suboptimal outcomes.

Nevertheless, what factors should be considered when determining $\tau$? To investigate this question, we revisit the wBC-based policy improvement method employed in IQL, as presented in Eq.~\eqref{Eq-IQL-Pi}. Formally, the objective of policy improvement can be interpreted as solving the following two-stage optimization problem,

\noindent First stage:
\begin{gather} 
\begin{aligned}
\pi^{*}(a|s) & = \mathop{\arg\max}_{\pi} \mathbb{E}_{a \sim \pi(\cdot|s)} \left[ A^{\pi_{\beta}}(s, a)\right], \\
\text{s.t.} \; & \mathop{D_{KL}} \left[\pi(\cdot|s)||\pi_{\beta}(\cdot|s) \right] \leq \epsilon, \\
& \int_{a} \pi(a|s) da = 1.
\label{Eq-opt-problem-1}
\end{aligned}
\end{gather}
\noindent Second stage:
\begin{gather} 
\begin{aligned}
\pi(\phi) = \mathop{\arg\min}_{\phi} \mathbb{E}_{\rho_{\pi_{\beta}}(s)} \left[ \mathop{D_{\text{KL}}} \left[ \pi^{*}(\cdot|s)||\pi_{\phi}(\cdot|s) \right]\right],
\label{Eq-opt-problem-2}
\end{aligned}
\end{gather}
where $\pi^{*}$ denotes the closed-form optimal solution, $\pi_\phi$ is the parametric policy to approximate $\pi^{*}$ by minimizing the KL divergence, and $\rho_{\pi_{\beta}}(s)$ signifies the state distribution in dataset.

According to the constraints in optimization problem \eqref{Eq-opt-problem-1} and \eqref{Eq-opt-problem-2}, the parameter $\tau$ should first be designed to reflect the distance $D(\pi_\phi||\pi_{\beta})$ between the learned policy $\pi_\phi$ and the behavior policy $\pi_{\beta}$. As $\pi_\phi$ approaches $\pi_{\beta}$, $\tau$ should increase optimistically; conversely, $\tau$ should decrease conservatively as they diverge.

Vector projection provides a way to identify the component of a vector along the direction of another vector. Specifically, the vector projection of vector $\vec{u}$ onto $\vec{v}$ is defined as follows,
$$
\text{Proj}_{\vec{v}}\vec{u} = \frac{\vec{u} \cdot \vec{v}}{\|\vec{v}\|} \frac{\vec{v}}{\|\vec{v}\|} = \frac{\vec{u} \cdot \vec{v}}{\|\vec{v}\|^2} \vec{v},
$$
where $\|\vec{x}\| = \sqrt{\sum_{i=1}^{m}x_i^2}$ denotes the $l_2$-norm of the vector $\vec{x} \in \mathbb{R}^m$. 

In practice, the learned policy $\pi_{\phi}$ is approximated by neural networks and represented by a $K$-dimensional diagonal Gaussian distribution, where $K$ represents the dimension of the action space. Typically, the policy networks are trained and updated using batch samples. Given $n$ input samples $\left[(s_1, a_1), (s_2, a_2), ... ,(s_n, a_n)\right]$, the policy network outputs $\vec{\pi}_{\phi}(a|s) = \left[\pi_{\phi}(a_1|s_1), \pi_{\phi}(a_2|s_2), ..., \pi_{\phi}(a_n|s_n)\right]$. Similarly, the behavior policy $\pi_\beta$ can also be approximated by a neural network, and output $\vec{\pi}_{\beta}(a|s) = \left[\pi_{\beta}(a_1|s_1), \pi_{\beta}(a_2|s_2), ..., \pi_{\beta}(a_n|s_n)\right]$.

Therefore, we propose parameter $\vec{\tau}_{\text{proj}}(a|s)$ based on the vector projection $\text{Proj}_{\vec{\pi}_{\phi}} \vec{\pi}_{\beta}$ of behavior policy vector $\vec{\pi}_{\beta}(a|s)$ onto the learned policy vector $\vec{\pi}_{\phi}(a|s)$, 
\begin{gather} 
\begin{aligned}
\vec{\tau}_{\text{proj}}(a|s) = \frac{\vec{\pi}_{\beta}(a|s) \cdot \vec{\pi}_{\phi}(a|s)}{\| \vec{\pi}_{\phi}(a|s) \|^2} \vec{\pi}_{\phi}(a|s).
\label{Eq-tau-proj}
\end{aligned}
\end{gather}

For one state-action pair in a batch sample, i.e. each component $\tau_{\text{proj}}(a|s)$ of the vector $\vec{\tau}_{\text{proj}}(a|s)$ is defined as follows,
\begin{gather} 
\begin{aligned}
\tau_{\text{proj}}(a|s) = \frac{\vec{\pi}_{\beta}(a|s) \cdot \vec{\pi}_{\phi}(a|s)}{\| \vec{\pi}_{\phi}(a|s) \|^2} \pi_{\phi}(a|s).
\label{Eq-tau-proj}
\end{aligned}
\end{gather}

Based on the $\tau_{\text{proj}}(a|s)$, the value function and $Q$-function for policy evaluation in PIQL are optimized according to the following loss functions,
{\small
\begin{gather} 
\begin{aligned}
L_V(\psi) = \mathop{\mathbb{E}}_{(s,a) \sim \mathcal{D}}\left[ L_2^{\tau_{\text{proj}}}(Q_{\hat{\theta}}(s,a) - V_\psi(s)) \right],
\label{Eq-PROJ-IQL-V}
\end{aligned}
\end{gather}
\begin{gather} 
\begin{aligned}
L_Q(\theta) = \mathop{\mathbb{E}}_{(s,a,s') \sim \mathcal{D}}\left[ (r(s,a) + \gamma V_\psi(s') - Q_{\theta}(s,a))^2 \right].
\label{Eq-PROJ-IQL-Q}
\end{aligned}
\end{gather}
}

The advantages of $\tau_{\text{proj}}(a|s)$ for policy evaluation can be summarized as follows: 

\emph{1. Reflecting Distance and Adaptively Adjust.} 

$\tau_{\text{proj}}(a|s)$ represents the length of the shadow of $\vec{\pi}_{\beta}(a|s)$ over $\vec{\pi}_{\phi}(a|s)$ at state-action pair $(s, a)$. As $\vec{\pi}_{\phi}(a|s)$ gets closer to $\vec{\pi}_{\beta}(a|s)$, the larger the projection, the larger $\tau_{\text{proj}}(a|s)$, thus allowing for an optimistic evaluation toward the policy $\pi_{\phi}(a|s)$; conversely, the small $\tau_{\text{proj}}(a|s)$ allows for a conservative evaluation. The trend of $\tau_{\text{proj}}(a|s)$ is consistent with the optimization problem \eqref{Eq-opt-problem-1} and \eqref{Eq-opt-problem-2}. Moreover, $\tau_{\text{proj}}(a|s)$ is calculated based on the batch sample under the current policy, not a pre-set hyperparameter. Hence, $\tau_{\text{proj}}(a|s)$ can also adaptively adjust without the requirement for time-consuming fine-tuning.

\emph{2. Multi-step yet In-sample Learning Paradigm.}

Based on the objective in Eq.~\eqref{Eq-IQL-Pi}, the actions for policy evaluation are drawn from the behavior policy $a \sim \pi_{\beta}$ in $f(s; a; \pi_{\beta})$, and the actions for policy improvement are from the dataset $a \sim \mathcal{D}$. Therefore, IQL is essentially a one-step algorithm. By contrast, PIQL can be classified as a multi-step algorithm, as the $\tau_{\text{proj}}$ parameter of value function Eq.~\eqref{Eq-PROJ-IQL-V} is determined based on information from the policy $\pi_{\phi}$. Furthermore, PIQL retains the expectile regression framework and the in-sample learning paradigm, effectively mitigating the risks associated with OOD actions. This property is formally established in Theorem~\ref{theorem1}.

\begin{theorem} \label{theorem1}
If the $\epsilon$ in the constraint Eq.~\eqref{Eq-opt-problem-1} is sufficiently small for $\{a | \pi_{\beta}(a|s)>0, \pi_{\phi}(a|s)>0 \text{, and } Q_{\hat{\theta}}(s, a) - V_\psi(s) < 0, \text{ for all } s\}$, Eq.~\eqref{Eq-PROJ-IQL-V} will be transformed as follows,
$$
L_V(\psi) = \mathop{\mathbb{E}}_{\substack{s \sim \mathcal{D}, \\ a \sim \pi_{\phi}(\cdot|s)}} \left[ L_2^{\Bar{\tau}_{\text{proj}}} (Q_{\hat{\theta}}(s,a) - V_\psi(s)) \right],
$$
where $\Bar{\tau}_{\text{proj}}(a|s) = \frac{\vec{\pi}_{\beta}(a|s) \cdot \vec{\pi}_{\phi}(a|s)}{\| \vec{\pi}_{\phi}(a|s) \|^2} \pi_{\beta}(a|s)$.
\end{theorem}

In Theorem~\ref{theorem1}, the loss function $L_{V}(\psi)$ is demonstrated to be equivalent to Eq.~\eqref{Eq-PROJ-IQL-V}. This equivalence implies that in Eq.~\eqref{Eq-PROJ-IQL-V}, the actions $a \sim \mathcal{D}$, when combined with the projection parameter $\tau_{\text{proj}}$, can be equivalently represented as $a \sim \pi_{\phi}$. Hence, PIQL can be interpreted as a multi-step algorithm. Furthermore, Theorem~\ref{theorem1} can be reformulated within a complete expectile regression framework by transforming $\tau_{\text{proj}}(a|s)$ into $\Bar{\tau}_{\text{proj}}(a|s)$. It should be emphasized that Theorem~\ref{theorem1} is derived under certain assumptions: the $\epsilon$ is sufficiently small for actions $\{a|Q_{\hat{\theta}}(s, a) - V_\psi(s) < 0\}$. These assumptions are reasonable, as actions $\{a|Q_{\hat{\theta}}(s, a) - V_\psi(s) < 0\}$ are generally regarded as suboptimal. For such actions, the optimal policy is to adopt imitation to avoid OOD overestimation, rather than engaging in unnecessary exploration of suboptimal actions.

\subsection{Policy Improvement in PIQL}

The policy improvement method in IQL is the wBC method, shown in Eq.~\eqref{Eq-IQL-Pi}. Nevertheless, directly applying the wBC approach to PIQL does not yield satisfactory results. This is because the policy struggles to converge to the optimal solution when $\epsilon$ is sufficiently small. Further details and theoretical justification are illuminated in the Lemma~\ref{lemma2}.

\begin{lemma} \label{lemma2}
If $\mathop{D_{KL}} \left[\pi_{\phi}(\cdot|s)||\pi_{\beta}(\cdot|s) \right] \leq \epsilon$, $\forall{s}$ is guaranteed, then the performance $\eta(\pi) = \mathbb{E}_{\tau \sim \pi} \left[ \sum_{t=0}^{\infty} \gamma^t r_t \right]$ has the following bound
$$
\eta(\pi_{\phi}) \leq \eta(\pi_{\beta}) + \frac{V_{\max}}{\sqrt{2}(1 - \gamma)} \sqrt{\epsilon},
$$
where $0 \leq Q^{\pi}, V^{\pi} \leq \frac{R_{\max}}{1-\gamma} =: V_{\max}$.
\end{lemma}

Based on the Lemma~\ref{lemma2}, to achieve the optimal policy with high probability, $\epsilon$ must be set to a large value. However, this requirement contradicts the assumption in Theorem~\ref{theorem1} of the policy evaluation stage, rendering the wBC policy improvement approach unsuitable for PIQL. By contrast, the support constraint-based policy improvement method, as presented in Eq.~\eqref{Eq-STR-Pi}, relaxes the density constraint. Consequently, we adopt the support constraint-based method for policy improvement in PIQL, shown as follows, 
\begin{gather} 
\begin{aligned}
J_\pi (\phi) & = \mathop{\max}_{\phi} \mathop{\mathbb{E}}_{\substack{s \sim \mathcal{D}, \\ a \sim \pi_{\phi}(\cdot|s)}} \left[ f(s, a; \Bar{\pi}_{\phi}, \Bar{\tau}_{\text{proj}}) \log \pi_\phi(a|s))\right] \\
& = \mathop{\max}_{\phi} \mathop{\mathbb{E}}_{(s,a) \sim \mathcal{D}} \left[ \frac{\Bar{\pi}_{\phi}(a|s)}{\pi_{\beta}(a|s)} f(s, a; \Bar{\pi}_{\phi}, \tau_{\text{proj}}) \log \pi_\phi(a|s) \right] \\
& = \mathop{\max}_{\phi} \mathop{\mathbb{E}}_{(s,a) \sim \mathcal{D}} \left[ \frac{\Bar{\pi}_{\phi}(a|s)}{\pi_{\beta}(a|s)} \exp\left(\frac{A^{\Bar{\pi}_{\phi}}_{\tau_{\text{proj}}}(s, a)}{\lambda} \right) \log \pi_\phi(a|s) \right],
\label{Eq-PROJ-IQL-PI} 
\end{aligned}
\end{gather}
where $A^{\Bar{\pi}_{\phi}}_{\tau_{\text{proj}}}(s, a) = Q^{\Bar{\pi}_{\phi}}_{\tau_{\text{proj}}}(s, a) - V^{\Bar{\pi}_{\phi}}_{\tau_{\text{proj}}}(s)$.

With an exact $Q$-function, we demonstrate that PIQL guarantees strict policy improvement, as demonstrated in Theorem~\ref{theorem2},

\begin{theorem} \label{theorem2}
If we have exact $Q$-function and $\tau_{k+1}(a|s) \geq \tau_{k}(a|s)$ \footnote{To simplify notation and without ambiguity, we denote the $\tau_{\text{proj}}(a|s)$ at the $k$-th iteration as $\tau_{k}(a|s)$.}, then $\pi_k$ in PIQL enjoys monotonic improvement:
$$
    Q^{\pi_{k+1}}_{\tau_{k+1}} (s, a) \geq Q^{\pi_{k}}_{\tau_{k}} (s, a), \quad \forall{s, a}.
$$
\end{theorem}

Theorem~\ref{theorem2} establishes that PIQL ensures monotonic policy improvement when policy optimization is performed under support constraint.

Compared with STR \cite{STR}, which also adopts a support-constraint approach, PIQL exhibits a clear advantage by enforcing a more rigorous criterion for evaluating \emph{advantageous actions}, where advantageous actions are defined as $\{a|Q(s, a) - V(s) \geq 0\}$. Intuitively, progressively tightening the advantageous actions evaluation criterion leads to a more ``concentrated'' policy, i.e., one that selects advantageous actions with higher probability.

To illustrate this advantage, we utilize the Monte Carlo estimates approach. Specifically, we sample a large number of actions according to a certain policy $a \sim \pi(\cdot|s)$ and compare the probabilities associated with advantageous actions. If
\begin{gather}
\begin{aligned}
P\left\{Q^{\pi_{k+1}}_{\tau_{k+1}}(s,a) -V^{\pi_{k+1}}_{\tau_{k+1}}(s) \geq 0\right\} \leq P\left\{Q^{\pi_{k}}_{\tau_{k}}(s,a) - V^{\pi_{k}}_{\tau_{k}}(s) \geq 0\right\},
\label{Eq-Criteria}
\end{aligned}
\end{gather}
Eq.~\eqref{Eq-Criteria} holds, then we define the evaluation criterion for advantageous actions as being more rigorous at $k+1$-th iteration, where $P\{\mathcal{M}\}$ denotes the probability of event $\mathcal{M}$.

We demonstrate that the expectation form of this more rigorous evaluation criterion is satisfied in Theorem~\ref{theorem3}.

\begin{theorem} \label{theorem3}
(Expectation Rigorous Criterion for Advantageous Actions). For $\forall s \in \mathcal{D}$ and $a \sim \pi_{k+1}(\cdot|s)$, if $0.5 \leq \tau_{k}(a|s), \tau_{k+1}(a|s) \leq 1$, we have
\begin{align*}
    \mathop{\mathbb{E}}_{\substack{a \sim \pi_{k+1}(\cdot|s)}} \left[  Q^{\pi_{k+1}}_{\tau_{k+1}}(s, a) - V^{\pi_{k+1}}_{\tau_{k+1}}(s) \right] \leq \mathop{\mathbb{E}}_{\substack{a \sim \pi_{k+1}(\cdot|s)}} \left[ Q^{\pi_{k}}_{\tau_{k}}(s,a) - V^{\pi_{k}}_{\tau_{k}}(s) \right].
\end{align*}
\end{theorem}

Theorem~\ref{theorem3} establishes the expectation form of the advantageous action evaluation criterion. Specifically, it states that the expectation of advantage function $A^{\pi_{k+1}}_{\tau_{k+1}}(s, a)$ under the policy $\pi_{k+1}$ is lower than the expectation of $A^{\pi_{k}}_{\tau_{k}}(s, a)$. Intuitively, for the same sampling policy $\pi_{k+1}$, if $A^{\pi_{k+1}}_{\tau_{k+1}}(s, a)$ is smaller than $A^{\pi_{k}}_{\tau_{k}}(s, a)$, then the corresponding expectation is also smaller. This result indicates that the evaluation criterion for advantageous actions at the $(k+1)$-th iteration, $A^{\pi_{k+1}}_{\tau_{k+1}}(s, a)$, is more rigorous than the previous criterion $A^{\pi_{k}}_{\tau_{k}}(s, a)$, as formalized in Eq.~\eqref{Eq-Criteria}.


Furthermore, Theorem~\ref{theorem4} extends result in Theorem~\ref{theorem3} to the maximum case of more rigorous evaluation criterion for two arbitrary policies. Before presenting Theorem~\ref{theorem4}, we first introduce Lemma~\ref{lemma3} as follows,

\begin{lemma} \label{lemma3}
For any random variable $X$, if $0.5 \leq \tau_1 \leq \tau_2 \leq 1$, we get
$$
    Var^{\tau_1}(X) \leq Var^{\tau_2}(X),
$$
where $Var^{\tau}(X) = \mathbb{E}[(X-\mathbb{E}^{\tau}(X))^2]$.
\end{lemma}

\begin{theorem} \label{theorem4}
(Maximum Rigorous Criterion for Advantageous Actions). For $\forall s \in \mathcal{D}$, if $0.5 \leq \tau_{k}(a|s) \leq \tau_{k+1}(a|s) \leq 1$, we have
\begin{align*}
\max_{a \sim \pi_1(\cdot|s)} P\left\{Q^{\pi_{k+1}}_{\tau_{k+1}}  (s, a)  - \right. & \left. V^{\pi_{k+1}}_{\tau_{k+1}}(s) \geq 0\right\} \leq \\ & \max_{a \sim \pi_2(\cdot|s)} P \left\{Q^{\pi_{k}}_{\tau_{k}}(s,a) - V^{\pi_{k}}_{\tau_{k}}(s) \geq 0\right\},
\end{align*}
where $\pi_1(\cdot|s)$ and $\pi_2(\cdot|s)$ are arbitrary policies.
\end{theorem}

Theorem~\ref{theorem4} generalizes Theorem~\ref{theorem3} by extending the sampling policy $\pi_{k+1}$ to two arbitrary policies. Intuitively, Theorem~\ref{theorem4} shows that the maximum probability of selecting advantageous actions becomes progressively more rigorous across iterations, regardless of which policies are utilized for sampling. This result indicates that even under the most favorable conditions, the evaluation criterion tightens over time, thereby ensuring that the policy improvement process consistently focuses on truly advantageous actions. In addition, Theorem~\ref{theorem4} is established under the assumption that $\tau_{k}(a|s) \leq \tau_{k+1}(a|s)$. In subsequent experiments, $\tau_{\text{proj}}(a|s)$ exhibits an increasing trend, which supports the validity of this assumption. 


\subsection{Practical Implementation in PIQL}

\textbf{Behavior Policy.} The behavior policy $\pi_{\beta}(a|s)$ is approximated by a neural network trained using the behavior cloning method, as defined in Eq.~\eqref{Eq-behavior-clone},
\begin{gather} 
\begin{aligned}
\pi_\beta(a|s) = \mathop{\text{argmax}}_{\pi} \mathbb{E}_{(s, a) \sim \mathcal{D}} \left[ \log \pi(a|s) \right].
\label{Eq-behavior-clone}
\end{aligned}
\end{gather}

\textbf{Parameter Processing.} In both Theorem~\ref{theorem3} and Theorem~\ref{theorem4}, we assume that the projection parameter satisfies $0.5 \leq \tau_{\text{proj}}(a|s) \leq 1$. In practice, this constraint is enforced using the \emph{clip} function, defined as
$$
    \tau_{\text{proj}}(a|s) = clip(\tau_{\text{proj}}(a|s), 0.5, 1),
$$
where $clip(x,l,u) = \min(\max(x,l),u)$.

In implementation, we observed that computing $\tau_{\text{proj}}(a|s)$ for each individual state–action pair leads to results that are overly fine-grained, thereby reducing the stability of policy evaluation. To mitigate this issue, we instead compute the mean value of $\tau_{\text{proj}}(a|s)$ across batch samples, which significantly improves stability.

\textbf{Importance Sampling.} To mitigate high variance as well as training instability, we employ the Self-Normalized Importance Sampling (SNIS) method \cite{STR}, which normalizes the IS ratios across the batch.

\textbf{Overall Architecture.} Integrating the policy evaluation, policy improvement, and practical implementation details, the complete PIQL algorithm is summarized in Algorithm~\ref{algorithm}.

\begin{algorithm}[!t]
\caption{PIQL}
\label{algorithm}
\hspace{-4cm} \textbf{Input}: Dataset $\mathcal{D} = \{(s, a, s', r)\}.$ \\
\textbf{Parameter}: $Q$-network $\theta$, target $Q$-network $\hat{\theta}$, value network $\psi$, policy network $\phi$. 
\begin{algorithmic}[1] 
\STATE {\textbf{\emph{// Pre-train behavior policy for $N_\beta$ steps.}}}
\FOR{$t=1$ to $N_\beta$}
\STATE Optimize the behavior policy by Eq.~\eqref{Eq-behavior-clone}.
\ENDFOR
\STATE {\textbf{\emph{// Policy Evaluation and Improvement.}}}
\FOR{each gradient step}
\STATE Random Batch Sample $(s, a, s', r) \sim \mathcal{D}$.
\STATE Calculate $\tau_{\text{proj}}(a|s)$ by Eq.~\eqref{Eq-tau-proj}.
\STATE Calculate batch mean value of $\tau_{\text{proj}}(a|s)$.
\STATE Optimize value network $\psi$ by Eq.~\eqref{Eq-PROJ-IQL-V}.
\STATE Optimize $Q$-network $\theta$ by Eq.~\eqref{Eq-PROJ-IQL-Q}.
\STATE Optimize policy network $\phi$ by Eq.~\eqref{Eq-PROJ-IQL-PI}.
\STATE Soft-update target Q-network $\theta' \leftarrow (1-\tau){\theta'} + \tau\theta$.
\ENDFOR
\STATE \textbf{return} Policy $\pi_\phi$.
\end{algorithmic}
\end{algorithm}

\section{Experiments} \label{Sec-Experiments}

\subsection{Comparisons on D4RL Benchmarks}

\begin{table*}[!t]
\caption{Averaged normalized scores on Gym-MuJoCo-v2 datasets. We indicate in bold that the scores are the top-$2$.}
\label{Tab-Gym}
\centering
\resizebox{1.0\linewidth}{!}{
\begin{tabular}{l||rrrrrrrrr|r} 
\toprule
Dataset                       & DT    & RvS   & POR       & CQL  & LAPO  & TD3+BC & Reinformer & CRR & IQL   & PIQL        \\
\hline
halfcheetah-med         & 42.6  & 41.6  & \textbf{48.8}      & 44.4  & 46.0  & 42.8   & 42.9 & 47.1 & 47.4  & \textbf{48.7}$\pm$0.45   \\
hopper-med              & 67.6  & 60.2  & 78.6      & \textbf{86.6}  & 51.6  & \textbf{99.5} & 81.6 & 38.1  & 66.3  & 68.2$\pm$13.90  \\
walker2d-med            & 74.0  & 71.7  & \textbf{81.1}      & 74.5  & 80.8  & 79.7   & 80.5 & 59.7 & 78.3  & \textbf{83.4}$\pm$0.63   \\
halfcheetah-med-rep  & 36.6  & 38.0  & 43.5      & \textbf{46.2}  & 43.9  & 43.3   & 39.0 & 44.4 & 44.2  & \textbf{45.5}$\pm$0.84   \\
hopper-med-rep       & 82.7  & 73.5  & \textbf{98.9}      & 48.6  & 37.6  & 31.4   & 83.3 & 25.5 & 94.7  & \textbf{95.7}$\pm$10.63  \\
walker2d-med-rep     & 66.6  & 60.6  & \textbf{76.6}      & 32.6  & 52.3  & 25.2   & 72.8 & 27.0 & 73.9  & \textbf{89.5}$\pm$2.68   \\
halfcheetah-med-exp  & 86.8  & 92.2  & \textbf{94.7}      & 62.4  & 86.1  & \textbf{97.9}  & 92.0 & 85.2  & 86.7  & 94.2$\pm$1.04   \\
hopper-med-exp       & 107.6 & 101.7 & 90.0      & \textbf{111.0} & 100.7  & \textbf{112.2}  & 107.8 & 53.0 & 91.5  & 101.5$\pm$19.82 \\
walker2d-med-exp     & 108.1 & 106.0 & 109.1     & 98.7  & 109.4  & 101.1  & 109.3   & 91.3 & \textbf{109.6} & \textbf{112.2}$\pm$0.93  \\
\hline
Gym-MuJoCo-v2 total           & 672.6 & 645.5 & \textbf{721.3}     & 605.0 & 608.4 & 633.1  & 709.2  & 471.3 & 692.4 & \textbf{738.9}$\pm$50.92 \\
\bottomrule
\end{tabular}
}
\end{table*}

Gym-MuJoCo Locomotion Tasks are popular benchmarks used in prior offline RL works. Accordingly, we compare PIQL with several competitive and commonly adopted baselines, including Decision Transformer (DT) \cite{DT}, RvS \cite{RvS}, POR \cite{POR}, CQL, LAPO \cite{LAPO}, TD3+BC \cite{TD3+BC}, Reinformer \cite{Reinformer}, CRR \cite{CRR} and IQL \cite{IQL}. The results are reported in Tab.~\ref{Tab-Gym}, where we average the mean return over $10$ evaluation trajectories and $5$ random seeds, and additionally report the standard deviation.

Although the performance of various offline RL algorithms on the Gym-MuJoCo-v2 datasets has largely saturated, PIQL continues to exhibit strong results, achieving top-$2$ performance on $6$ out of $9$ tasks and obtaining the highest overall score across all $9$ datasets, thereby demonstrating the effectiveness of the proposed approach.

AntMaze and Kitchen Navigation tasks present significantly greater challenges, compared to the Gym-MuJoCo locomotion tasks. The main challenge is to learn policies for long-horizon planning from datasets that do not contain optimal trajectories. We evaluate PIQL against several competitive methods, including BC, BCQ \cite{BCQ}, BEAR, CQL, Onestep, TD3+BC, Diffusion-QL \cite{Diffusion-QL}, $\mathcal{X}$-QL \cite{XQL}, and IQL, utilizing the same dataset versions as employed in IQL for a fair comparison. The results are presented in Tab.~\ref{Tab-Ant-Kit}.

\begin{table*}[!t]
\caption{Averaged normalized scores on  AntMaze-v0, and Kitchen-v0 datasets. We indicate in bold that the scores are the top-$2$.}
\label{Tab-Ant-Kit}
\centering
\resizebox{1.0\linewidth}{!}{
\begin{tabular}{l||rrrrrrrrr|r}
\toprule
Dataset                       & BC    & BCQ   & BEAR     & CQL   & Onestep  & TD3+BC & Diffusion-QL & $\mathcal{X}$-QL  & IQL   & PIQL        \\
\hline
antmaze-umaze              & 54.6  & 78.9  & 73.0     & 74.0  & 62.4     & 78.6   & 93.4         & \textbf{93.8}     & 87.5  & \textbf{98.0}$\pm$6.0 \\
antmaze-umaze-diverse      & 45.6  & 55.0  & 61.0     & \textbf{84.0}  & 44.7     & 71.4   & 66.2         & 82.0              & 62.2  & \textbf{98.0}$\pm$6.0 \\
antmaze-med-play        & 0.0   & 0.0   & 0.0      & 61.2  & 5.4      & 10.6   & \textbf{76.6}         & 76.0              & 71.2  & \textbf{94.0}$\pm$14.0 \\
antmaze-med-diverse     & 0.0   & 0.0   & 8.0      & 53.7  & 1.8      & 3.0    & \textbf{78.6}         & 73.6              & 70.0  & \textbf{88.0}$\pm$8.0 \\
antmaze-large-play         & 0.0   & 6.7   & 0.0      & 15.8  & 0.1      & 0.2    & 46.4         & \textbf{46.5}              & 39.6  & \textbf{72.0}$\pm$9.8 \\
antmaze-large-diverse      & 0.0   & 2.2   & 0.0      & 14.9  & 0.9      & 0.0    & \textbf{56.6}         & 49.0              & 47.5  & \textbf{50.0}$\pm$10.0 \\
\hline
AntMaze-v0 total              & 100.2 & 142.8 & 142.0    & 303.6 & 115.2    & 163.8  & 417.8        & \textbf{420.9}             & 378.0 & \textbf{500.0}$\pm$53.8 \\
\hline
kitchen-complete           & 65.0  & 8.1   & 0.0      & 43.8  & 66.0     & 61.5   & \textbf{84.0}         & 82.4              & 62.5  & \textbf{87.0}$\pm$4.6 \\
kitchen-partial            & 38.0  & 18.9  & 13.1     & 49.8  & 59.3     & 52.8   & 60.5         & \textbf{73.7}              & 46.3  & \textbf{67.0}$\pm$6.0 \\
kitchen-mixed              & 51.5  & 8.1   & 47.2     & 51.0  & 56.5     & 60.8   & \textbf{62.6}         & \textbf{62.5}              & 51.0  & 55.5$\pm$3.2 \\
\hline
Kitchen-v0 total              & 154.5 & 35.1  & 60.3     & 144.6 & 181.7    & 175.1  & 207.1        & \textbf{218.6}             & 159.8 & \textbf{209.5}$\pm$13.8 \\
\bottomrule
\end{tabular}
}
\end{table*}

The Tab.~\ref{Tab-Ant-Kit} reveal that imitation learning methods such as BC and BCQ struggle to achieve satisfactory results. This limitation arises due to sparse rewards and a large amount of suboptimal trajectories, which place higher demands on algorithms for stable and robust $Q$-function and value function estimation. PIQL addresses these challenges by dynamically and adaptively tuning the $\tau_{\text{proj}}$ parameter to estimate the $Q$-function and value function according to the current policy. As a result, PIQL demonstrates exceptional performance on the AntMaze-v0 and Kitchen-v0 datasets, particularly excelling in AntMaze, where it achieves the highest scores on $5$ out of the $6$ datasets.

\subsection{The Training Curves of $\tau_\text{proj}$}

\begin{figure*}[!t] 
\centerline{\includegraphics[width=0.92\textwidth]{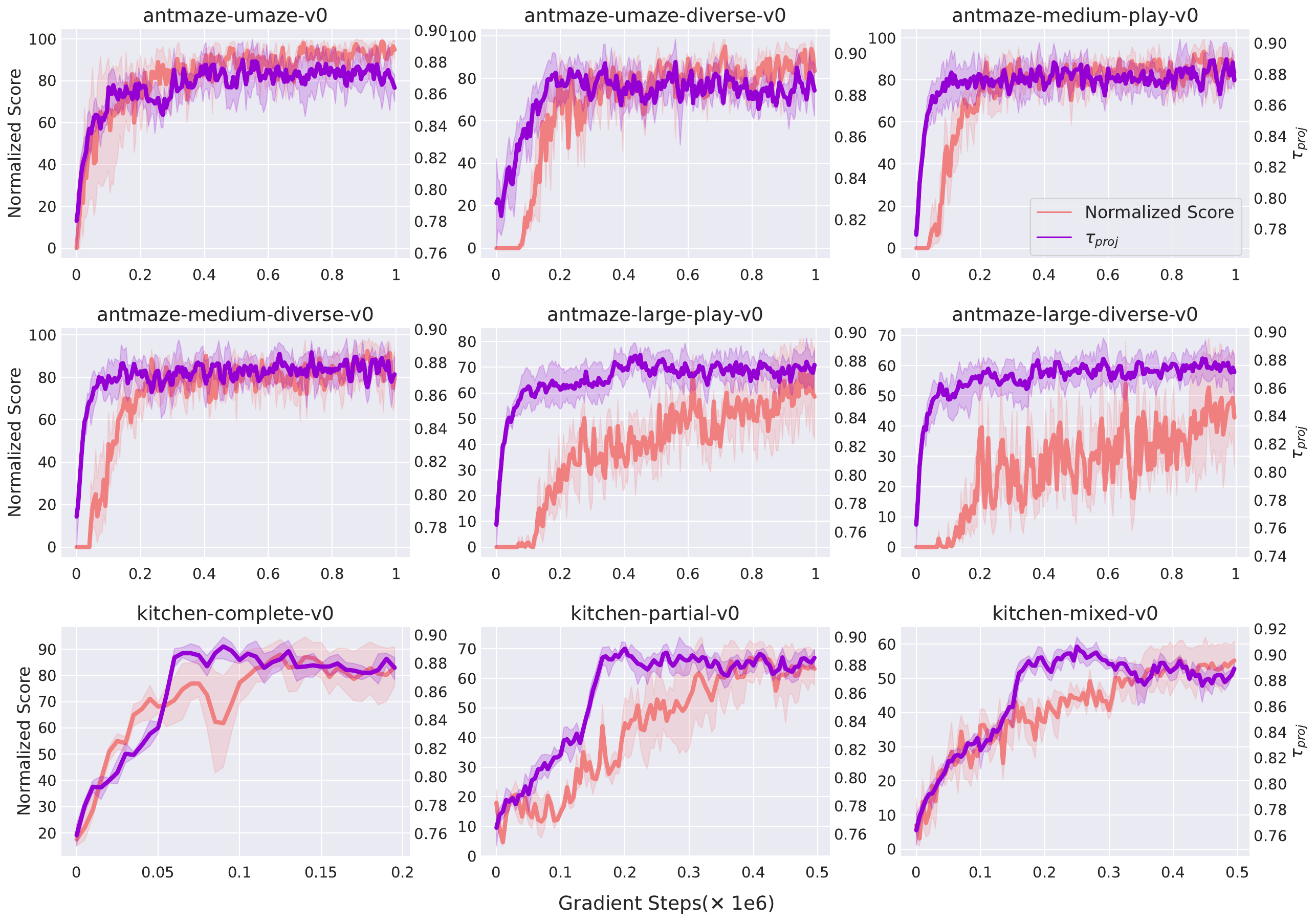}}
\caption{The training curves of normalized score and $\tau_{\text{proj}}(a|s)$ on AntMaze-v0 and Kitchen-v0 datasets.}
\label{Fig-weight-score}
\end{figure*}

In contrast to the fixed hyperparameter $\tau$ in IQL, $\tau_\text{proj}(a|s)$ can be flexibly and adaptively tuned. Therefore, we plot the $\tau_\text{proj}(a|s)$ curve and normalized score curve on AntMaze-v0 and Kitchen-v0 datasets. The results are shown in Fig.~\ref{Fig-weight-score}.

Based on Fig.~\ref{Fig-weight-score}, we observed that the curve $\tau_{\text{proj}}(a|s)$ exhibits an upward trend, reflecting an increase in the projected component of $\pi_\phi(a|s)$ within $\pi_\beta(a|s)$. This phenomenon arises because, in PIQL, the updated method of policy can be seen as a weighted behavior cloning as described in Eq.~\eqref{Eq-PROJ-IQL-PI}, where the weight factor is given by $\frac{\Bar{\pi}_{\phi}(a|s)}{\pi_{\beta}(a|s)} \exp\Big( \frac{A^{\Bar{\pi}_{\phi}}_{\tau_{\text{proj}}}(s, a)}{\lambda} \Big) > 0$. However, this weight is essentially different from the weight in the aforementioned wBC method, since the support constraint is constructed via importance sampling. Furthermore, this upward trend supports our assumption $0.5 \leq \tau_{k}(a|s) \leq \tau_{k+1}(a|s) \leq 1$ in Theorem~\ref{theorem2} and Theorem~\ref{theorem4}.

Another noteworthy observation is that the rise in normalized scores lags behind the $\tau_{\text{proj}}(a|s)$ parameter, indicating that updates to $\tau_{\text{proj}}(a|s)$ contribute to more accurate estimates of the $Q$-function and value function. Then, the more accurate estimates facilitate continuous policy improvement, resulting in higher normalized scores. However, this phenomenon is less apparent in the kitchen-complete-v0 dataset, since this dataset is relatively small with only $3,680$ samples. In contrast, datasets such as AntMaze-v0 typically contain approximately $10^6$ samples, and other Kitchen-v0 datasets include $136,950$ samples.

\begin{table*}[!t]
\caption{Averaged normalized scores on NeoRL2 datasets. We indicate in bold that the scores are the top-$2$.}
\label{Tab-NeoRL2}
\centering
\resizebox{1.0\linewidth}{!}{
\begin{tabular}{l||rrrrrrrrrr|r}
\toprule
Dataset               & Data   & BC   & CQL   & EDAC  & MCQ   & TD3+BC & MOPO  & COMBO & RAMBO   & MOBILE & PIQL \\
\hline
Pipeline              & 69.3  & 68.6  & 81.1  & 72.9  & 49.7  & \textbf{82.0}   & -26.3 & 55.5  & 24.1    & 65.5   & \textbf{89.3}$\pm$14.8\\
Simglucose            & 73.9  & \textbf{75.1}  & 11.0  & 8.1   & 29.6  & 74.2   & 34.6  & 23.2  & 10.8    & 9.3    & \textbf{82.1}$\pm$3.6\\
RocketRecovery        & 75.3  & 72.8  & 74.3  & 65.7  & 76.5  & \textbf{79.7}   & -27.7 & 74.7  & -44.2   & 43.7   & \textbf{81.6}$\pm$7.5\\
RandomFrictionHopper  & 28.7  & 28.0  & 33.0  & \textbf{34.7}  & 31.7  & 29.5   & 32.5  & 34.1  & 29.6    & \textbf{35.1}   & 34.2$\pm$0.7\\
DMSD                  & 56.6  & 65.1  & 70.2  & \textbf{78.7}  & \textbf{77.8}  & 60.0   & 68.2  & 68.3  & 76.2    & 64.4   & 54.2$\pm$0.1\\
Fusion                & 48.8  & 55.2  & 55.9  & 58.0  & 49.7  & 54.6   & -11.6 & 55.5  & \textbf{59.6}    & 5.0    & \textbf{59.8}$\pm$2.1\\
SafetyHalfCheetah     & \textbf{73.6}  & 70.2  & \textbf{71.2}  & 53.1  & 54.7  & 68.6   & 23.7  & 57.8  & -422.4  & 8.7    & 70.7$\pm$1.2\\
\hline
NeoRL2 total          & 426.2 & 435.0 & 396.7 & 371.2 & 369.7 & \textbf{448.6}  & 93.4  & 369.1 & -266.3  & 231.7  & \textbf{471.9} $\pm$ 30.0\\
\bottomrule
\end{tabular}}
\end{table*}

\subsection{Ablation Experiment}

\begin{figure*}[!t] 
\centerline{\includegraphics[width=0.92\textwidth]{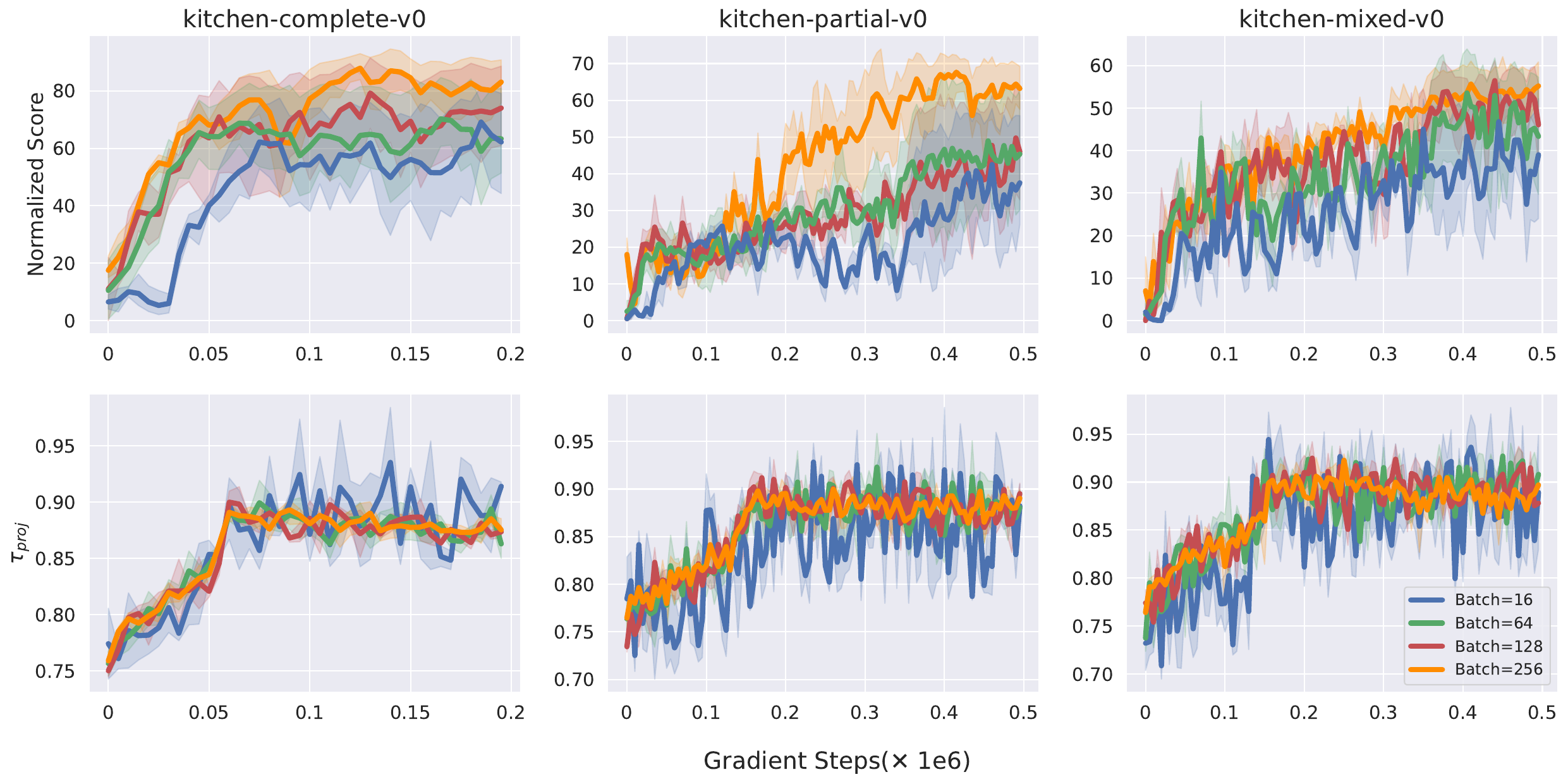}}
\caption{The training curves of normalized score and $\tau_{\text{proj}}(a|s)$ on Kitchen-v0 datasets under different batch size.}
\label{Fig-Batch}
\end{figure*}

In the proposed PIQL method, each sample in the batch is treated as a component of the vector used to compute $\tau_{\text{proj}}(a|s)$, making the \emph{batch size} to be one of the critical parameters. To investigate this, we evaluated the performance of $\tau_{\text{proj}}(a|s)$ as well as normalized scores corresponding $batch \; size = 16, 64, 128, 256$ on three Kitchen-v0 datasets. The results are presented in Fig.~\ref{Fig-Batch}.

As shown in Fig.~\ref{Fig-Batch}, the normalized score gradually increases with larger batch sizes, which aligns with expectations. This improvement occurs because larger batch sizes allow both the policy and value networks to be trained more effectively. However, this trend does not fully capture the specific impact of batch size on the parameter $\tau_{\text{proj}}(a|s)$. To address this issue, we also plotted the relationship between batch size and $\tau_{\text{proj}}(a|s)$. The results clearly indicate that the stability of $\tau_{\text{proj}}(a|s)$ improves as the batch size increases, with reduced fluctuations. A more stable $\tau_{\text{proj}}(a|s)$ is particularly beneficial for enhancing performance because of the policy evaluation is more stable. 


\subsection{Comparisons on NeoRL2 Benchmarks}

NeoRL2 (Near Real-World Offline RL Benchmark) \cite{NeoRL2} is a recent benchmark designed to reflect practical constraints in offline RL, which includes domains motivated by industrial control,  nuclear fusion, and healthcare. Moreover, practical data acquisition limits the dataset size and coverage.

We compare PIQL against baselines reported on NeoRL2, and the results are shown in Tab.~\ref{Tab-NeoRL2}. Based on Tab.~\ref{Tab-NeoRL2}, PIQL attains the best performance on $4$ out of $7$ datasets and yields the highest total score, substantially outperforming the strongest baseline.

\section{Conclusion} \label{Sec-Conclusion}
We propose PIQL, an in-sample, multi-step, support-constraint offline RL algorithm based on IQL. During the policy evaluation phase, PIQL transitions from a one-step approach to a multi-step in-sample offline RL algorithm while preserving the expectile regression framework. In the policy improvement phase, PIQL incorporates a support constraint to facilitate effective policy updates. Furthermore, we provide a theoretical foundation, demonstrating the invariance of the expectile regression framework, monotonic policy improvement guarantees, and a more rigorous policy evaluation criterion. Empirical results validate the theoretical insights of PIQL and highlight its state-of-the-art performance on D4RL benchmarks. In the future, first, we will study how to make PIQL more data-efficient, aiming to learn strong policies from substantially smaller datasets. Second, we will explore replacing the behavior policy with more expressive models, aiming to further enhance robustness and generalization in practical applications.


\bibliographystyle{ACM-Reference-Format} 
\bibliography{sample}


\newpage
\appendix
\onecolumn

\centerline{\huge\textbf{Appendix}}

\section{Proofs}

\noindent \textbf{Proof of Lemma~\ref{lemma1}}.

Lemma \ref{lemma1}. \emph{For all $s$, $\tau_1$ and $\tau_2$ such that $\tau_1 < \tau_2$ we get
$$
V_{\tau_1}(s) \leq V_{\tau_2}(s).
$$}
\begin{proof} \label{proof-Lemma1}
The proof follows the \emph{Lemma.2} proof in \cite{IQL}.
\end{proof}

\vspace{3mm}
\noindent \textbf{Proof of Theorem~\ref{theorem1}}.

Theorem~\ref{theorem1}. \emph{If the $\epsilon$ in the constraint Eq.~\eqref{Eq-opt-problem-1} is sufficiently small for all $a \in \{ \pi_{\beta}(a|s)>0, \pi_{\phi}(a|s)>0 \text{, and } Q_{\hat{\theta}}(s, a) - V_\psi(s) < 0, \text{ for all } s.\}$,  then Eq.~\eqref{Eq-PROJ-IQL-V} will be transformed as follows,
$$
L_V(\psi) = \mathop{\mathbb{E}}_{\substack{s \sim \mathcal{D}, \\ a \sim \pi_{\phi}(\cdot|s)}} \left[ L_2^{\Bar{\tau}_{\text{proj}}} (Q_{\hat{\theta}}(s,a) - V_\psi(s)) \right].
$$
where $\Bar{\tau}_{\text{proj}}(a|s) = \frac{\vec{\pi}_{\beta}(a|s) \cdot \vec{\pi}_{\phi}(a|s)}{\| \vec{\pi}_{\phi}(a|s) \|^2} \pi_{\beta}(a|s)$.}

\begin{proof}
For one state-action pair $(s, a)$, the $\tau_{\text{proj}}(a|s)$ is defined as Eq.~\eqref{Eq-tau-proj}. We bring the $\tau_{\text{proj}}(a|s)$ into Eq.~\eqref{Eq-PROJ-IQL-V} and expand,
\begin{gather} 
\begin{aligned}
L_V(\psi) & = \mathop{\mathbb{E}}_{(s,a) \sim \mathcal{D}}\left[ L_2^{\tau_{\text{proj}}}(Q_{\hat{\theta}}(s,a) - V_\psi(s)) \right] \\ 
& = \begin{cases}
     \mathop{\mathbb{E}}_{(s,a) \sim \mathcal{D}} \left[\tau_{\text{proj}} (Q_{\hat{\theta}}(s,a) - V_\psi(s))^2\right], & \text{if } Q_{\hat{\theta}}(s,a) - V_\psi(s) \geq 0; \\
     \mathop{\mathbb{E}}_{(s,a) \sim \mathcal{D}} \left[(1 - \tau_{\text{proj}}) (Q_{\hat{\theta}}(s,a) - V_\psi(s))^2\right], & \text{if } Q_{\hat{\theta}}(s,a) - V_\psi(s) < 0.
\end{cases}
\label{Eq-PROJ-IQL-Expand}
\end{aligned}
\end{gather}

\textcircled{1}. For $Q_{\hat{\theta}}(s,a) - V_\psi(s) \geq 0$,
\begin{gather} 
\begin{aligned}
L_V^{Q \geq V}(\psi) & = \mathop{\mathbb{E}}_{(s,a) \sim \mathcal{D}} \left[\tau_{\text{proj}} (Q_{\hat{\theta}}(s,a) - V_\psi(s))^2\right] \\
& = \mathop{\mathbb{E}}_{\substack{s \sim \mathcal{D}, \\ a \sim \pi_{\beta}(\cdot|s)}} \left[ \frac{\vec{\pi}_{\beta}(a|s) \cdot \vec{\pi}_{\phi}(a|s)}{\| \vec{\pi}_{\phi}(a|s) \|^2} \pi_{\phi}(a|s) (Q_{\hat{\theta}}(s,a) - V_\psi(s))^2 \right] \\
& = \mathop{\mathbb{E}}_{\substack{s \sim \mathcal{D}, \\ a \sim \pi_{\phi}(\cdot|s)}} \left[ \frac{\vec{\pi}_{\beta}(a|s) \cdot \vec{\pi}_{\phi}(a|s)}{\| \vec{\pi}_{\phi}(a|s) \|^2} \pi_{\beta}(a|s) (Q_{\hat{\theta}}(s,a) - V_\psi(s))^2 \right].
\label{Eq-PROJ-IQL-Q>V} 
\end{aligned}
\end{gather}
 
\textcircled{2}. For $Q_{\hat{\theta}}(s,a) - V_\psi(s) < 0$,
\begin{gather} 
\begin{aligned}
L_V^{Q<V}(\psi) & = \mathop{\mathbb{E}}_{(s,a) \sim \mathcal{D}} \left[ (1 - \tau_{\text{proj}}) (Q_{\hat{\theta}}(s,a) - V_\psi(s))^2\right] \\ 
& = \mathop{\mathbb{E}}_{\substack{s \sim \mathcal{D}, \\ a \sim \pi_{\beta}(\cdot|s)}} \left[ \left(1 - \frac{\vec{\pi}_{\beta}(a|s) \cdot \vec{\pi}_{\phi}(a|s)}{\| \vec{\pi}_{\phi}(a|s) \|^2} \pi_{\phi}(a|s) \right) (Q_{\hat{\theta}}(s,a) - V_\psi(s))^2 \right] \\
& = \mathop{\mathbb{E}}_{\substack{s \sim \mathcal{D}, \\ a \sim \pi_{\beta}(\cdot|s)}}\left[ (Q_{\hat{\theta}}(s,a) - V_\psi(s))^2 \right] - L_V^{Q \geq V}(\psi).
\label{Eq-PROJ-IQL-Q<V} 
\end{aligned}
\end{gather}

In addition, we calculate
\begin{gather} 
\begin{aligned}
& \left| \mathop{\mathbb{E}}_{\substack{s \sim \mathcal{D}, \\ a \sim \pi_{\phi}(\cdot|s)}}\left[ (Q_{\hat{\theta}}(s,a) - V_\psi(s))^2 \right] - \mathop{\mathbb{E}}_{\substack{s \sim \mathcal{D}, \\ a \sim \pi_{\beta}(\cdot|s)}}\left[ (Q_{\hat{\theta}}(s,a) - V_\psi(s))^2 \right] \right| \\
& = \left| \mathop{\mathbb{E}}_{s \sim \mathcal{D}} \left[ \int (\pi_{\phi}(a|s) - \pi_{\beta}(a|s))(Q_{\hat{\theta}}(s,a) - V_\psi(s))^2 da \right] \right| \\
& \leq \mathop{\mathbb{E}}_{s \sim \mathcal{D}} \left[ \int \left| \pi_{\phi}(a|s) - \pi_{\beta}(a|s)\right| (Q_{\hat{\theta}}(s,a) - V_\psi(s))^2 da \right] \\
& \leq \mathop{\mathbb{E}}_{s \sim \mathcal{D}} \left[ \int D_{TV}(\pi_{\phi}(\cdot|s)\|\pi_{\beta}(\cdot|s)) (Q_{\hat{\theta}}(s,a) - V_\psi(s))^2 da \right] \\
& \leq \frac{1}{\sqrt{2}} \mathop{\mathbb{E}}_{s \sim \mathcal{D}} \left[ \int \sqrt{D_{KL}(\pi_{\phi}(\cdot|s)\|\pi_{\beta}(\cdot|s))} (Q_{\hat{\theta}}(s,a) - V_\psi(s))^2 da \right] \qquad (\text{Pinsker's Inequality}) \\
& \leq \sqrt{\frac{\epsilon}{2}} \mathop{\mathbb{E}}_{s \sim \mathcal{D}} \left[ \int (Q_{\hat{\theta}}(s,a) - V_\psi(s))^2 da \right].
\label{Eq-Two-Pi-Q-V}
\end{aligned}
\end{gather}

Under the assumption that the $\epsilon$ in the constraint of Eq.~\eqref{Eq-opt-problem-1} is sufficiently small, we have the difference between $\mathop{\mathbb{E}}_{s \sim \mathcal{D}, a \sim \pi_{\phi}(\cdot|s)}\left[ (Q_{\hat{\theta}}(s, a) - V_\psi(s))^2 \right]$ and $\mathop{\mathbb{E}}_{s \sim \mathcal{D}, a \sim \pi_{\beta}(\cdot|s)}\left[ (Q_{\hat{\theta}}(s,a) - V_\psi(s))^2 \right]$ is enough small based on the Eq.~\eqref{Eq-Two-Pi-Q-V}.

Therefore, for $Q_{\hat{\theta}}(s,a) - V_\psi(s) < 0$, we can rewrite the Eq.~\eqref{Eq-PROJ-IQL-Q<V} as follow,
\begin{gather} 
\begin{aligned}
L_V^{Q<V}(\psi) & = \mathop{\mathbb{E}}_{\substack{s \sim \mathcal{D}, \\ a \sim \pi_{\beta}(\cdot|s)}}\left[ (Q_{\hat{\theta}}(s,a) - V_\psi(s))^2 \right] - L_V^{Q \geq V}(\psi) \\ 
&  \approx \mathop{\mathbb{E}}_{\substack{s \sim \mathcal{D}, \\ a \sim \pi_{\phi}(\cdot|s)}}\left[ (Q_{\hat{\theta}}(s,a) - V_\psi(s))^2 \right] - L_V^{Q \geq V}(\psi) \\
& = \mathop{\mathbb{E}}_{\substack{s \sim \mathcal{D}, \\ a \sim \pi_{\phi}(\cdot|s)}}\left[ \left(1 - \frac{\vec{\pi}_{\beta}(a|s) \cdot \vec{\pi}_{\phi}(a|s)}{\| \vec{\pi}_{\phi}(a|s) \|^2} \pi_{\beta}(a|s)\right) (Q_{\hat{\theta}}(s,a) - V_\psi(s))^2\right].
\label{Eq-PROJ-IQL-Q<V-Re} 
\end{aligned}
\end{gather}

Combining Eq.~\eqref{Eq-PROJ-IQL-Q>V} and \eqref{Eq-PROJ-IQL-Q<V-Re}, we obtain
\begin{gather} 
\begin{aligned}
L_V(\psi) = \mathop{\mathbb{E}}_{\substack{s \sim \mathcal{D}, \\ a \sim \pi_{\phi}(\cdot|s)}} \left[ L_2^{\Bar{\tau}_{\text{proj}}} (Q_{\hat{\theta}}(s,a) - V_\psi(s)) \right].
\label{Eq-PROJ-IQL-New}
\end{aligned}
\end{gather}

where $\Bar{\tau}_{\text{proj}}(a|s) = \frac{\vec{\pi}_{\beta}(a|s) \cdot \vec{\pi}_{\phi}(a|s)}{\| \vec{\pi}_{\phi}(a|s) \|^2} \pi_{\beta}(a|s)$.
\end{proof}

\vspace{3mm}
\noindent \textbf{Proof of Lemma~\ref{lemma2}}.

Lemma \ref{lemma2}. \emph{If $\mathop{D_{KL}} \left[\pi_{\phi}(\cdot|s)||\pi_{\beta}(\cdot|s) \right] \leq \epsilon$, $\forall{s}$ is guaranteed, then the performance $\eta(\pi) = \mathbb{E}_{\tau \sim \pi} \left[ \sum_{t=0}^{\infty} \gamma^t r_t \right]$ has the following bound
$$
\eta(\pi_{\phi}) \leq \eta(\pi_{\beta}) + \frac{V_{\max}}{\sqrt{2}(1 - \gamma)} \sqrt{\epsilon}.
$$
where $0 \leq Q^{\pi}, V^{\pi} \leq \frac{R_{\max}}{1-\gamma} =: V_{\max}$.}
\begin{proof}
    The proof follows the \emph{Lemma 3.1} proof in \cite{STR}.
\end{proof}

\vspace{3mm}
\noindent \textbf{Proof of Theorem~\ref{theorem2}}.

Theorem \ref{theorem2}. \emph{If we have exact $Q$-function and $\tau_{k+1}(a|s) \geq \tau_{k}(a|s)$, then $\pi_k$ in Proj-IQL enjoys monotonic improvement:
$$
    Q^{\pi_{k+1}}_{\tau_{k+1}} (s, a) \geq Q^{\pi_{k}}_{\tau_{k}} (s, a), \quad \forall{s, a}.
$$}
\begin{proof}
The $\pi_{\phi}(a|s)$ under the policy improvement objective Eq.~\eqref{Eq-PROJ-IQL-PI} is a parametric approximation of 
\begin{gather} 
\begin{aligned}
    \pi_{k+1}(a|s) = \frac{1}{Z_{k}(s)} \pi_{k}(a|s) \exp \left( \frac{A^{\pi_{k}}_{\tau_k}(s,a)}{\lambda} \right).
\label{Eq-PROJ-IQL-PI-ORI} 
\end{aligned}
\end{gather}
where $Z_{k}(s) = \int_{a} \pi_{k}(a|s) \exp\Big(\frac{A^{\pi_{k}}_{\tau_k}(s,a)}{\lambda} \Big) da$, and $A^{\pi_{k}}_{\tau_k}(s,a) = Q^{\pi_{k}}_{\tau_k}(s,a) - V^{\pi_{k}}_{\tau_k}(s)$.

Based on the wBC policy extraction methods, the Eq.~\eqref{Eq-PROJ-IQL-PI-ORI} is the optimal solution for the following optimization problem,
\begin{gather} 
\begin{aligned}
\pi_{k+1} & = \mathop{\arg\max}_{\pi \in \Pi} \mathbb{E}_{a \sim \pi(\cdot|s)} \left[ Q^{\pi_k}_{\tau_{k}} (s,a)\right], \\
\text{s.t.} \; & \mathop{D_{KL}} \left[\pi(\cdot|s)||\pi_{k}(\cdot|s) \right] \leq \epsilon, \\
& \int_{a} \pi(a|s) da = 1.
\label{Eq.24 Eq-opt-problem-PROJ}
\end{aligned}
\end{gather}

Because $\mathop{D_{KL}}[\pi_{k}(\cdot|s)\|\pi_{k}(\cdot|s)] =0 < \epsilon, ~ \forall s$ is a strictly feasible solution. Therefore, ${\mathbb{E}}_{a \sim \pi_{k+1}} [Q^{\pi_{k}}_{\tau_k} (s,a)] \geq {\mathbb{E}}_{a \sim \pi_{k}} [Q^{\pi_{k}}_{\tau_k} (s,a)], ~\forall s$. It implies
\begin{align*}
& Q^{\pi_{k}}_{\tau_k} (s,a) \\
& = r(s, a) + \gamma \mathbb{E}_{s_{t+1} \sim p(\cdot|s_t, a_t)} \mathbb{E}_{a \sim \pi_{k}} ^ {\tau_k} \left[ Q^{\pi_{k}}_{\tau_k} (s_{t+1}, a_{t+1}) | s_t=s, a_t=a \right] \\
& \leq r(s, a) + \gamma \mathbb{E}_{s_{t+1} \sim p(\cdot|s_t, a_t)} \mathbb{E}_{a \sim \pi_{k+1}} ^ {\tau_k} \left[ Q^{\pi_{k}}_{\tau_k} (s_{t+1}, a_{t+1}) | s_t=s, a_t=a \right] \\
&\quad\quad\quad\quad\quad \ldots \\
& \leq \mathbb{E}_{a \sim \pi_{k+1}} ^ {\tau_k} \left[ \sum_{n=0}^{\infty} \gamma^{n} r(s_{t+n}, a_{t+n})| s_t=s, a_t=a \right]\\
& = Q^{\pi_{k+1}}_{\tau_k}(s, a).
\end{align*}

Therefore, $Q^{\pi_{k+1}}_{\tau_k}(s, a) \geq Q^{\pi_{k}}_{\tau_k}(s, a), ~ \forall s,a$. 

Then, we rewrite $Q^{\pi_{k+1}}_{\tau_k}(s, a)$ as 
\begin{align*}
& Q^{\pi_{k+1}}_{\tau_k}(s, a) \\
& = r(s, a) + \gamma \mathbb{E}_{s_{t+1} \sim p(\cdot |s_t, a_t)} \left[V^{\pi_{k+1}}_{\tau_k}(s_{t+1}) | s_t=s, a_t=a \right] \\
& \leq r(s, a) + \gamma \mathbb{E}_{s_{t+1} \sim p(\cdot |s_t, a_t)} \left[V^{\pi_{k+1}}_{\tau_{k+1}}(s_{t+1}) | s_t=s, a_t=a \right] \qquad (Lemma.~\ref{lemma1}) \\
& = Q^{\pi_{k+1}}_{\tau_{k+1}}(s, a).
\end{align*}

Therefore, $Q^{\pi_{k+1}}_{\tau_{k+1}}(s, a) \geq Q^{\pi_{k+1}}_{\tau_{k}}(s, a) \geq  Q^{\pi_{k}}_{\tau_k}(s, a), ~ \forall s,a$
\end{proof}

\vspace{3mm}
\noindent \textbf{Proof of Theorem~\ref{theorem3}}.

Theorem \ref{theorem3}. \emph{(Expectation Rigorous Criterion for Superior Actions). For $\forall s \in \mathcal{D}$ and $a \sim \pi_{k+1}(\cdot|s)$, if $0.5 \leq \tau_{k}(a|s), \tau_{k+1}(a|s) \leq 1$, we have
\begin{align*}
    \mathbb{E}_{a \sim \pi_{k+1}(\cdot|s)} \left[  Q^{\pi_{k+1}}_{\tau_{k+1}}(s, a)  -  V^{\pi_{k+1}}_{\tau_{k+1}}(s) \right] \leq \mathbb{E}_{a \sim \pi_{k+1}(\cdot|s)} \left[ Q^{\pi_{k}}_{\tau_{k}}(s,a) - V^{\pi_{k}}_{\tau_{k}}(s) \right].
\end{align*}}
\begin{proof}

For the left-hand side of the above inequality,
\begin{gather}  
\begin{aligned}
& \mathbb{E}_{a \sim \pi_{k+1}(\cdot|s)} \left[ Q^{\pi_{k+1}}_{\tau_{k+1}}(s,a) - V^{\pi_{k+1}}_{\tau_{k+1}}(s) \right] \\
& = \mathbb{E}_{a \sim \pi_{k+1}(\cdot|s)} \left[ Q^{\pi_{k+1}}_{\tau_{k+1}}(s,a)\right] - V^{\pi_{k+1}}_{\tau_{k+1}}(s) \\
& = V^{\pi_{k+1}}(s) - V^{\pi_{k+1}}_{\tau_{k+1}}(s) \\
& \leq 0. \qquad (Lemma.~\ref{lemma1})
\end{aligned}
\end{gather}

For the right-hand side of the inequality,
\begin{gather}  
\begin{aligned}
& \mathbb{E}_{a \sim \pi_{k+1}(\cdot|s)} \left[ Q^{\pi_{k}}_{\tau_{k}}(s,a) - V^{\pi_{k}}_{\tau_{k}}(s) \right] \\
& = \mathbb{E}_{a \sim \pi_{k+1}(\cdot|s)} \left[ Q^{\pi_{k}}_{\tau_{k}}(s,a) \right] - V^{\pi_{k}}_{\tau_{k}}(s) \\
& = \mathbb{E}_{a \sim \pi_{k}(\cdot|s)}^{\tau \rightarrow 1} \left[ Q^{\pi_{k}}_{\tau_{k}}(s,a) \right] - V^{\pi_{k}}_{\tau_{k}}(s) \qquad ( 
Eq.~\eqref{Eq.24 Eq-opt-problem-PROJ} ~ \& ~Theorem.~\ref{theorem2}) \\
& = V^{\pi_{k}}_{\tau \rightarrow 1}(s) - V^{\pi_{k}}_{\tau_{k}}(s) \\
& \geq 0. \qquad  (Lemma.~\ref{lemma1})
\end{aligned}
\end{gather}

Therefore, Theorem.~\ref{theorem3} holds.
\end{proof}

\vspace{3mm}
\noindent \textbf{Proof of Lemma~\ref{lemma3}}.

Lemma \ref{lemma3}. \emph{For any random variable $X$, if $0.5 \leq \tau_1 \leq \tau_2 \leq 1$, we get
$$
    Var^{\tau_1}(X) \leq Var^{\tau_2}(X).    
$$
where $Var^{\tau}(X) = \mathbb{E}[(X-\mathbb{E}^{\tau}(X))^2]$.}
\begin{proof}
We calculate
\begin{gather} %
\begin{aligned}
& Var^{\tau_1}(X) - Var^{\tau_2}(X) \\
& = \mathbb{E}(X^2) - 2\mathbb{E}(X)\mathbb{E}^{\tau_1}(X) + \mathbb{E}^{\tau_1}(X)^2 - \mathbb{E}(X^2) + 2\mathbb{E}(X)\mathbb{E}^{\tau_2}(X) - \mathbb{E}^{\tau_2}(X)^2 \\
& = \underbrace{\left[ \mathbb{E}^{\tau_1}(X) + \mathbb{E}^{\tau_2}(X) - 2 \mathbb{E}(X) \right]}_{\geq 0}  \underbrace{\left[ \mathbb{E}^{\tau_1}(X) - \mathbb{E}^{\tau_2}(X) \right]}_{\leq 0} \qquad (Lemma.\ref{lemma1} \; and \; 0.5 \leq \tau_1 \leq \tau_2 \leq 1) \\
& \leq 0.
\nonumber
\end{aligned}
\end{gather}

Therefore, $Var^{\tau_1}(X) - Var^{\tau_2}(X) \leq 0$.
\end{proof}

\vspace{3mm}
\noindent \textbf{Proof of Theorem~\ref{theorem4}}.

Theorem \ref{theorem4}. \emph{(Maximum Rigorous Criterion for Superior Actions). For $\forall s \in \mathcal{D}$, if $0.5 \leq \tau_{k}(a|s) \leq \tau_{k+1}(a|s) \leq 1$, we have
\begin{align*}
\max_{a \sim \pi_1(\cdot|s)} P\left\{Q^{\pi_{k+1}}_{\tau_{k+1}} (s, a)  - V^{\pi_{k+1}}_{\tau_{k+1}}(s) \geq 0\right\} \leq \max_{a \sim \pi_2(\cdot|s)} P \left\{Q^{\pi_{k}}_{\tau_{k}}(s,a) - V^{\pi_{k}}_{\tau_{k}}(s) \geq 0\right\}.
\end{align*}
where $\pi_1(\cdot|s)$ and $\pi_2(\cdot|s)$ are arbitrary policies.}
\begin{proof}
Firstly, we define 
\begin{gather} 
\begin{aligned} 
\mathbb{E}_{a \sim \pi_{k+1}}^{\tau_{k+1}}\left[ Q^{\pi_{k+1}}_{\tau_{k+1}}(s,a) \right] - \mathbb{E}_{a \sim \pi_{k+1}} \left[ Q^{\pi_{k+1}}_{\tau_{k+1}}(s,a) \right] = \epsilon_1.
\label{Eq-def-epsilon1}
\end{aligned}
\end{gather}
\begin{gather} 
\begin{aligned} 
\mathbb{E}_{a \sim \pi_{k}}^{\tau_{k}}\left[ Q^{\pi_{k}}_{\tau_{k}}(s,a) \right] - \mathbb{E}_{a \sim \pi_{k}} \left[ Q^{\pi_{k}}_{\tau_{k}}(s,a) \right] = \epsilon_2.
\label{Eq-def-epsilon2}
\end{aligned}
\end{gather}
Because $0.5 \leq \tau_{k}(a|s) \leq \tau_{k+1}(a|s) \leq 1$ and Lemma.~\ref{lemma1}, $\epsilon_1, \epsilon_2 \geq 0$.

Next, we found that for any random variable $X$,
\begin{gather} 
\begin{aligned} 
& Var^{\tau}(X) - Var(X) \\
& = \mathbb{E}(X)^2 - 2\mathbb{E}(X)\mathbb{E}^{\tau}(X) + \mathbb{E}^{\tau}(X)^2 \\
& = \left[ \mathbb{E}(X) - \mathbb{E}^{\tau}(X) \right] ^2 \\
& \geq 0.
\label{Eq-var-difference}
\end{aligned}
\end{gather}

Based on Eq.~\eqref{Eq-var-difference} and the definition in Eq.~\eqref{Eq-def-epsilon1} and \eqref{Eq-def-epsilon2}, we get
\begin{gather} 
\begin{aligned} 
Var^{\tau_{k+1}}\left[ Q^{\pi_{k+1}}_{\tau_{k+1}}(s,a) \right] - Var\left[ Q^{\pi_{k+1}}_{\tau_{k+1}}(s,a) \right] = \epsilon_1^2.
\label{Eq-var-difference1}
\end{aligned}
\end{gather}
\begin{gather} 
\begin{aligned} 
Var^{\tau_{k}}\left[ Q^{\pi_{k}}_{\tau_{k}}(s,a) \right] - Var\left[ Q^{\pi_{k}}_{\tau_{k}}(s,a) \right] = \epsilon_2^2.
\label{Eq-var-difference2}
\end{aligned}
\end{gather}

Then, based on the Cantelli’s inequality (one-sided Chebyshev’s inequality),
\begin{gather} 
\begin{aligned} 
& P\{Q^{\pi_{k+1}}_{\tau_{k+1}}(s,a) - V^{\pi_{k+1}}_{\tau_{k+1}}(s) \geq 0\} \\
& = P\left\{Q^{\pi_{k+1}}_{\tau_{k+1}}(s,a) - \mathbb{E}^{\tau_{k+1}}_{a \sim \pi_{k+1}} \left[ Q^{\pi_{k+1}}_{\tau_{k+1}}(s,a) \right] \geq 0 \right\} \\
& = P\left\{Q^{\pi_{k+1}}_{\tau_{k+1}}(s,a) - \mathbb{E}_{a \sim \pi_{k+1}} \left[ Q^{\pi_{k+1}}_{\tau_{k+1}}(s,a) \right] \geq \epsilon_1 \right\} \qquad (Eq.~\eqref{Eq-def-epsilon1}) \\
& \leq \frac{Var\left[ Q^{\pi_{k+1}}_{\tau_{k+1}}(s,a) \right]}{Var\left[ Q^{\pi_{k+1}}_{\tau_{k+1}}(s,a) \right] + \epsilon_1^2} \qquad (Cantelli’s \; Inequality)\\
& = \frac{Var\left[ Q^{\pi_{k+1}}_{\tau_{k+1}}(s,a) \right]}{Var^{\tau_{k+1}}\left[ Q^{\pi_{k+1}}_{\tau_{k+1}}(s,a)\right]}.
\label{Eq-inequality1}
\end{aligned}
\end{gather}

Similarly, 
\begin{gather} 
\begin{aligned} 
P\{Q^{\pi_{k}}_{\tau_{k}}(s,a) - V^{\pi_{k}}_{\tau_{k}}(s) \geq 0\} \leq \frac{Var\left[ Q^{\pi_{k}}_{\tau_{k}}(s,a) \right]}{Var^{\tau_{k}}\left[ Q^{\pi_{k}}_{\tau_{k}}(s,a)\right]}.
\label{Eq-inequality2}
\end{aligned}
\end{gather}

When $0.5 \leq \tau_{k}(a|s) \leq \tau_{k+1}(a|s) \leq 1$, we have $Var^{\tau_{k+1}}(X) \geq Var^{\tau_{k}}(X) \geq Var(X) \geq 0$. Therefore, combining Eq.~\eqref{Eq-inequality1} with \eqref{Eq-inequality2}, Theorem.~\ref{theorem4} holds.
\end{proof}

\section{Experimental Details} \label{Apppendix-B}
The vast majority of the hyperparameters in our experiments are consistent with those used in IQL, as summarized in Tab.~\ref{Tab-Hyperparameters}. 

\begin{table*}[t]
\caption{Hyperparameters}
\label{Tab-Hyperparameters}
\centering
\begin{tabular}{c|c|l}
\toprule
Hyperparameter & Value & Environment / Network \\
\hline
Hidden Layers             & 2                 & All \\
Layer Width               & 256               & All \\
Activations               & RuLU              & All \\
Learning Rate             & 3e-4              & All \\
Soft Update               & 5e-3              & All \\
\multirow{3}*{Training Gradient Steps}   & 1e6               & Gym-MuJoCo-v2 \& AntMaze-v0 tasks \\
                        ~ & 2e5               & Kitchen-complete-v0 task\\
                        ~ & 5e5               & Kitchen-partial-v0 \& Kitchen-mixed-v0 tasks\\
Evaluation Epochs         & 10                & All \\
Evaluation Frequency      & 5e3               & All \\
Random Seed               & [0, 2, 4, 6, 8]   & All \\
Batch Size                & 256               & All \\
\multirow{3}*{Inverse Temperature $\frac{1}{\lambda}$}& 10.0              & AntMaze-v0 \\
		            ~ & 3.0               & Gym-MuJoCo-v2 \\
                        ~ & 0.5               & Kitchen-v0 \\
Behavior Policy Training Step & 1e5          & All \\
\multirow{3}*{$\tau$}     & 0.9               & AntMaze-v0 \\
                        ~ & 0.7               & Gym-MuJoCo-v2 \\
                        ~ & 0.7               & Kitchen-v0 \\
Exponentiated Advantage   & $(-\infty, 100]$  & All \\
\multirow{4}*{Dropout}    & 0                 & All Q networks and target-Q networks \\
                        ~ & 0                 & All value networks \\
                        ~ & 0                 & Policy networks for AntMaze-v0 and Gym-MoJoCo-v2 tasks \\
                        ~ & 0.1               & Policy networks for Kitchen-v0 tasks \\

\bottomrule
\end{tabular}
\end{table*}

The hyperparameters specified for each experiment are described below. Any hyperparameters not explicitly mentioned will default to those shown in Tab.~\ref{Tab-Hyperparameters}.

\textbf{For the experiments corresponding to Fig.~\ref{Fig-different-tau}}. We adjusted only the $\tau = 0.3, 0.6, 0.9$. All other hyperparameters remained consistent with those listed in Tab.~\ref{Tab-Hyperparameters}.

\textbf{For the experiments corresponding to Fig.~\ref{Fig-weight-score}}. We referenced the parameters listed in Tab.~\ref{Tab-Hyperparameters}. In addition, following the suggestions of the authors of the dataset, we subtract $1$ from rewards for the AntMaze-v0 datasets.

\textbf{For the experiments corresponding to Fig.~\ref{Fig-Batch}}. We referenced the parameters listed in Tab.~\ref{Tab-Hyperparameters}, but the batch size is modified as $16, 64, 128, 256$.

\textbf{For the experiments corresponding to Tab.~\ref{Tab-Gym}, \ref{Tab-Ant-Kit} and \ref{Tab-NeoRL2}}. We referenced the parameters listed in Tab.~\ref{Tab-Hyperparameters}.

\end{document}